\documentclass[runningheads]{llncs}
\usepackage{multirow}
\usepackage{tabularx}
\usepackage{bbm}
\usepackage{adjustbox}
\usepackage{booktabs}
\newcolumntype{C}[1]{>{\centering\let\newline\\\arraybackslash\hspace{0pt}}m{#1}}
\newcolumntype{L}[1]{>{\raggedright\let\newline\\\arraybackslash\hspace{0pt}}m{#1}}

\usepackage[T1]{fontenc}
\usepackage{marvosym}
\usepackage{amsfonts}
\usepackage{amsmath}
\usepackage{graphicx,verbatim}
\usepackage{color}
\usepackage{tikz}
\usetikzlibrary{shapes}
\usepackage{hyperref}     

\begin{document}
\newcommand{\markerone}{\raisebox{0.5pt}{\tikz{\node[draw,scale=0.4,circle,fill=none](){};}}} 
\newcommand{\markertwo}{\raisebox{0.5pt}{\tikz{\node[draw,scale=0.4,regular polygon, regular polygon sides=4,fill=none](){};}}} 
\newcommand{\markerthree}{\raisebox{0.5pt}{\tikz{\node[draw,scale=0.3,regular polygon, regular polygon sides=3,fill=none,rotate=0](){};}}} 
\title{Learning to Stop: Reinforcement Learning for Efficient Patient-Level Echocardiographic Classification}
\titlerunning{Learning to Stop}
%
\authorrunning{Cho Kim et al.}
\author{Woo-Jin Cho Kim\inst{1}, Jorge Oliveira\inst{1}, Arian Beqiri\inst{1}, Alex Thorley\inst{1}, Jordan Strom\inst{2}, Jamie O'Driscoll\inst{3,4}, Rajan Sharma\inst{4}, Jeremy Slivnick\inst{5}, Roberto Lang\inst{5}, Alberto Gomez\inst{1}, Agisilaos Chartsias\inst{1,}\textsuperscript{\Letter}}
\institute{Ultromics Ltd., Oxford, UK, \textsuperscript{\Letter}\href{mailto:agis.chartsias@ultromics.com}{\email{agis.chartsias@ultromics.com}} \and
Beth Israel Deaconess Medical Center, Boston MA, USA \and University of Leicester, Leicester, UK \and 
St George’s University Hospitals NHS Foundation Trust, London, UK \and University of Chicago Medicine, Chicago IL, USA
}
%
%
\maketitle              

\begin{abstract}
Guidelines for transthoracic echocardiographic examination recommend the acquisition of multiple video clips from different views of the heart, resulting in a large number of clips. Typically, automated methods, for instance disease classifiers, either use one clip or average predictions from all clips. Relying on one clip ignores complementary information available from other clips, while using all clips is computationally expensive and may be prohibitive for clinical adoption.
%
To select the optimal subset of clips that maximize performance for a specific task (image-based disease classification), we propose a method optimized through reinforcement learning. In our method, an agent learns to either keep processing view-specific clips to reduce the disease classification uncertainty, or stop processing if the achieved classification confidence is sufficient. Furthermore, we propose a learnable attention-based aggregation method as a flexible way of fusing information from multiple clips. The proposed method obtains an AUC of 0.91 on the task of detecting cardiac amyloidosis using only 30\% of all clips, exceeding the performance achieved from using all clips and from other benchmarks.
%
%
\end{abstract}
\section{Introduction}

Echocardiography (echo) is widely used to support cardiac diagnosis
, but the quality of echo acquisition and interpretation depends on the sonographer's experience~\cite{MitchellCarol2019GfPa}. Deep learning has thus been increasingly explored to assist in disease detection~\cite{ehj,goto2021artificial,germain2023deep,zhang2023deep,peng2024deep,yu2023deep}. 
%
Most disease detection methods use a single echo clip~\cite{goto2021artificial,ehj,Akerman2023_HFpEF}, although guidelines recommend acquiring multiple clips per patient study, including variations of the same view (e.g. unfocused or focused on a ventricle)~\cite{MitchellCarol2019GfPa,robinson2020guide,lang2015recommendations}. Often, sonographers acquire the same view multiple times to ensure image quality. Therefore, automatic disease classification from a single clip requires clip selection, discarding information available in other clips. The naive solution of processing all clips is computationally expensive and may be prohibitive in time-constrained clinical workflows. 

Processing multiple (or all) clips in a study can be performed with Multiple Instance Learning (MIL) methods, which learn to aggregate multiple instances of a sample~\cite{carbonneau2018multiple}. Instance-based methods aggregate clip-level predictions using pooling operators~\cite{holste2022self,wessler2023automated} and were applied in echo for severity grade classification of aortic stenosis~\cite{pmlr-v219-huang23a}. Embedding-based methods aggregate latent feature vectors~\cite{pmlr-v219-huang23a}. 
Here, we aim to aggregate multiple clips, as in MIL, and to select the smallest number of clips from each study by considering a clip processing cost. 
%
Most cost-aware methods 
are designed to select a fixed set of features based on average costs observed during training or validation~\cite{ng2004feature,xu2014classifier}. Although effective in reducing computational burden, they lack adaptability as features are not adjusted per sample at inference time. 
%
To address this limitation, active acquisition methods acquire information sequentially based on what has already been observed~\cite{shim18_afa}. Reinforcement learning (RL) lends itself naturally to sequential decision making problems and has been used to optimize feature selection from different imaging modalities and minimize unnecessary acquisitions~\cite{bernardino2022reinforcement}, although it does not scale to an arbitrary number of instances per modality; in our case instances and modalities being analogous to clips and views, respectively. 

In this paper, we address the clip selection problem when classifying cardiac amyloidosis from echo studies with an arbitrary number of clips, acquired from three common views: apical four chamber (A4C), parasternal long-axis (PLAX), and parasternal short-axis (PSAX). 
To reduce the number of clips processed for each study, we propose a cost-sensitive, view-aware RL framework that handles a variable number of clips and selects the minimum required. 
Guided by a cost-accuracy reward, an agent trained via Proximal Policy Optimization (PPO) aggregates feature embeddings (generated from a pre-trained amyloidosis classifier) into a study-level representation using attention and decides whether to keep processing clips sequentially, which view to process next if so; or terminate. 

\textbf{Contributions:}
In summary, our contributions are:
(1) we propose a RL framework, agnostic to the size of echo studies, to optimize the selection of clips, improve disease classification performance and minimize processing costs;  
(2) we demonstrate the effectiveness of our framework for amyloidosis detection on an independent multi-site echo dataset. 



\section{Methods}
\begin{figure}[t!]
\centering
\includegraphics[width=\textwidth]{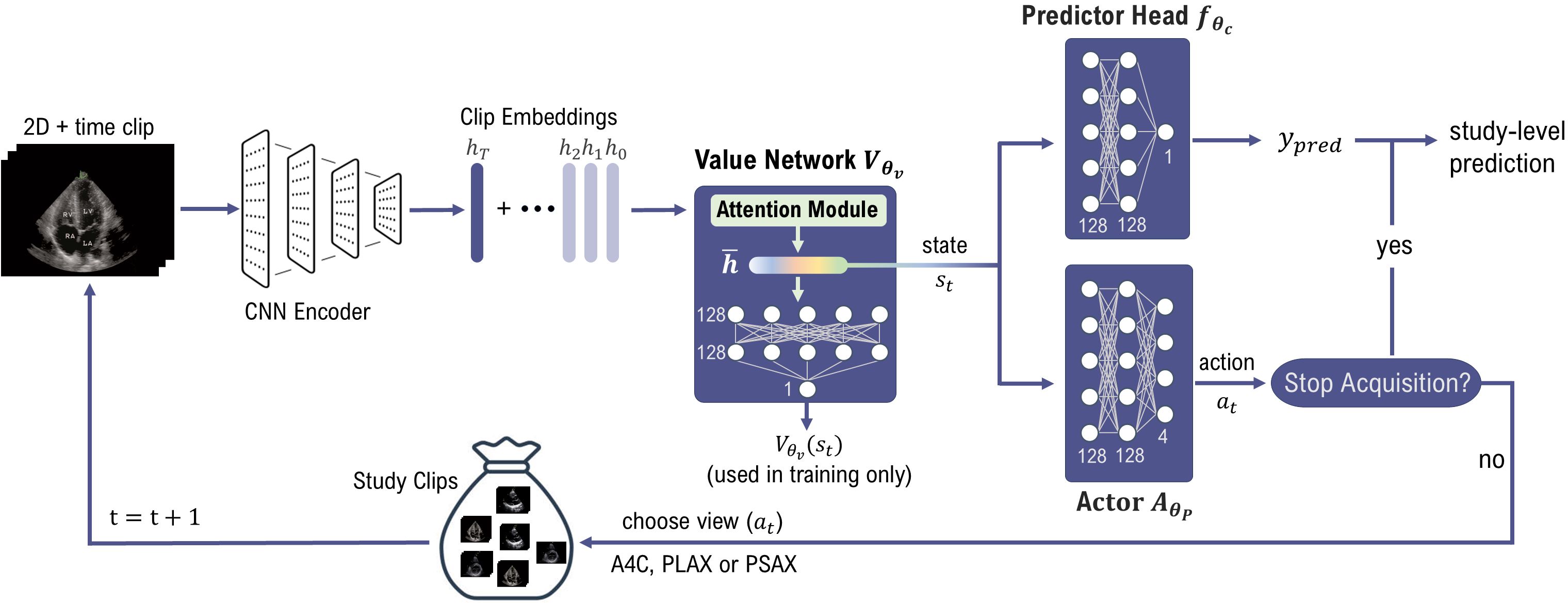}
\caption{Policy-guided sequential view selection for study-level prediction. At inference step $t$, a clip is encoded into an embedding $\mathbf{h}_t$ by a CNN encoder. The value network $V_{\theta_v}$ pools the embedding set \{$\mathbf{h}_1$,...,$\mathbf{h}_t$\} into a unified state $\mathbf{\bar{h}}=s_t$. The actor $A_{\theta_P}$ then chooses an action $a_t\in\{A4C, PLAX, PSAX, Stop\}$. If $Stop$, $y_{pred}$ becomes the final study-level prediction, if not, a new clip of view $a_t$ is sampled, restarting the process.}

\label{method_diagram}
\end{figure}

We frame amyloidosis detection as a sequential decision-making problem, where an agent decides whether to process an additional clip from a view or make the final prediction, if the information gathered is decided to be sufficient. We use an online RL framework, illustrated in Fig.~\ref{method_diagram}, where an agent dynamically selects actions that maximize a reward: a balance between accuracy and processing cost.

We first train a randomly initialized Convolutional Neural Network (CNN) for amyloidosis classification on a dataset comprising of studies with one or more clips from the A4C, PLAX, and PSAX views. All clips had the same disease label for a study/patient.
Given step $T$ where the agent (actor) has processed $T-1$ clips, clip $T$ is passed through the CNN encoder to produce an embedding $\mathbf{h}_T \in \mathbb{R}^{D}$. An attention module learns weights $\{ \mathbf{w}_t \in \mathbb{R}^D \}_{t=1}^T$
from all $\mathbf{h}$ vectors.
An element-wise softmax across the $T$ weight vectors yields attention weights $\mathbf{\beta}_t$, which pool the embeddings in a unified representation $\mathbf{\bar{h}}$, inspired by~\cite{pmlr-v219-huang23a}:
\begin{equation}
\mathbf{\bar{h}} = \sum_{t=0}^{T} \boldsymbol{\beta}_t \cdot \mathbf{h}_t, \qquad   \qquad \ \boldsymbol{\beta}_t(k) = \frac{e^{(\mathbf{w}_t(k))}}{\sum_{j=0}^{T} e^{(\mathbf{w}_j(k))}} ,
\label{eq_att}
\end{equation}
where $k$ indexes the $D$ embedding dimension, and $\cdot$ denotes element-wise multiplication. 
The unified embedding $\mathbf{\bar{h}}$ is passed to an actor $A_{\theta_P}$ that samples an action $a_t$ 
and an amyloidosis predictor $f_{\theta_C}$. The RL framework is detailed below.

%
\subsection{Reinforcement Learning}
Sequential clip selection is modeled as a Markov Decision Process (MDP), where an episode corresponds to the processing of a patient's study comprising multiple clips. The agent begins from state $s_0$ and progresses through a sequence of states, actions, and rewards respectively $(s_0, a_0, r_0), ..., (s_n, a_n, r_n)$. The state space $S$ consists of the set of clip embeddings $\{\mathbf{h}_1,...,\mathbf{h}_T\}$ generated by the frozen encoder of the pre-trained  CNN amyloidosis classifier. The action space $A$ is the set of possible actions: request a clip from a specific view $v \in \{A4C, PLAX, PSAX\}$ or terminate and make a prediction. The episode ends when the agent takes the stop action or when all available clips have been used. The reward function $R(s_t, a_t)$ encourages accurate predictions while penalizing unnecessary clip processing~\cite{bernardino2022reinforcement}. The reward for non-final states is 0, whereas for the final state $s_n$ is:
\begin{equation}
\begin{aligned}
R(s_n, a_n) &= b_y \cdot \mathbb{I} \big( y_{pred} = y_{gt} \big) - \lambda_{cost} \sum_{t=0}^n c_t, \quad & b_{y} &= \frac{1}{p(y_{gt})},
\end{aligned}
\end{equation}
where $\mathbb{I}$ is the indicator function that assigns a reward of $1$ if the prediction $y_{pred} = \mathbb{I}(f_{\theta_c}(\bar{h}) \geq 0.5)$ matches the ground truth $y_{gt}$ and $0$ otherwise. The term $b_y$ is a class-balancing weight that inversely scales with the class frequency $p(y_{gt})$, ensuring that under-represented classes contribute more to the learning signal~\cite{lin2020deep}. Term $c_t$ represents the cost of processing clip $t$\footnote{The processing cost $c_t$ is assumed constant, but can also be a function of the input.} and $\lambda_{cost}$ balances a trade-off between performance and processing cost.

Under this MDP, we learn with RL an optimal policy $\pi^*$ defined as a mapping $\pi: S\rightarrow A$. The optimal policy $\pi^*$  for a given state, is the sequence of actions that maximizes the expected discounted sum of rewards across all episodes of the training data: 
\begin{equation}
\pi^* = \arg\max_{\pi} \mathbb{E}_{\tau \sim \pi}\left[ \sum_{t=0}^{n} \gamma^t R(s_t, a_t) \right] = \arg\max_{\pi} \mathbb{E}_{\tau \sim \pi} \left[R_\tau \right],
\end{equation}
where $\tau = (s_0, a_0, \dots, s_n, a_n)$ is a trajectory generated by following policy $\pi$, and $\gamma$ is a discount factor that prioritizes immediate rewards. Since the transition and reward functions are unknown, RL estimates $\pi^*$ from sampled episodes. We adopt an actor-critic architecture~\cite{schulman2017proximal}, where a policy network $A_{\theta_P}(s_t)$ (\textit{actor}), with parameters $\theta_P$, receives a state and outputs a probability distribution over possible actions, and a value network $V_{\theta_V}(s_t)$ (\textit{critic}), with parameters $\theta_V$, estimates the expected discounted sum of rewards $R_t = \mathbb{E}\left[\textstyle\sum_{l=0}^{T - t} \gamma^l R(s_{t + l}, a_{t + l})\right]$. The value network is trained with a squared error loss:
\begin{equation}
\mathcal{L}_V (\theta_V) = \mathbb{E}_{\tau \sim \pi} \left[ (V_{\theta_V}(s_t) - R_t)^2 \right].
\label{eq_value_loss}
\end{equation}
\\
The policy network is optimized using PPO, with a clipped objective between $1-\epsilon$ and $1+\epsilon$ to prevent large updates~\cite{schulman2017proximal}:
\begin{equation}
\mathcal{L}_P (\theta_P) = \mathbb{E}_{s_t, a_t} \left[ \min( r_t(\theta_p) A_t, \;\text{clip}(r_t(\theta_p), 1 - \epsilon, 1 + \epsilon) A_t ) \right],
\end{equation}
where $r_t(\theta_P)=\textstyle A_{\theta_P}(s_t) / A_{\theta_{P_{\mathrm{old}}}}(s_t)$ is the ratio between the policy probabilities after and before parameter updates, and $\epsilon$ is the clip coefficient. For the policy update, we use General Advantage Estimation (GAE) to reduce variance in the gradient updates~\cite{schulman2015high}. The advantage function is computed as:
\begin{equation}
A_t = \sum_{l=0}^{\infty} (\gamma \,\lambda_{advantage})^l \,\delta_{t+l},
\quad
\delta_{t+l} = r_{t+l} + \gamma\,V_{\theta_V}(s_{t+l+1}) - V_{\theta_V}(s_{t+l}),
\end{equation}
where $\lambda_{advantage}$ is a decay-factor, $l$ is the look-ahead step, and $\delta_{t+l}$ is the temporal difference error at step $t+l$ . An entropy loss term is also included to encourage exploration and prevent premature policy convergence \cite{schulman2017proximal}. 
Furthermore, the predictor head $f_{\theta_c}$ is trained with binary cross-entropy for amyloidosis classification $\mathcal{L}_c(\theta_c)=-y_{gt}\log(y_{pred}) + (1 - y_{gt})\log(1 - y_{pred})$.
The final loss function combines policy, entropy, value, and classification losses:
\begin{equation}
\mathcal{L} (\theta_P, \theta_V, \theta_c) = \mathcal{L}_P(\theta_P) - \lambda_{\text{entropy}} \mathcal{L}_{\text{entropy} }(\theta_P) + \lambda_V \mathcal{L}_V(\theta_V) +
\mathcal{L}_c(\theta_c),
\end{equation}
where $\lambda_{\text{entropy}}$ and  $\lambda_V$ are hyperparameters to control the influence of entropy regularization and value function loss, respectively.

\subsection{Embedding pooling and networks setup}
The value network $V_{\theta_v}$ is a four-layer multi-layer perceptron (MLP) with an intermediate output of the attention module, which pools individual embeddings to a unified $\mathbf{\bar{h}}$ (Fig.~\ref{method_diagram}). The attention module is trained jointly via the value loss $\mathcal{L}_V$ and classification loss $\mathcal{L}_c$. To allow for a variable number of embeddings per study, we create a 2D matrix of dimensions $(D, N_{max})$, where $N_{max}$ is the maximum number of clips in a study.  Studies with fewer embeddings than $N_{max}$ are padded with zero-valued embeddings and the corresponding weights $\mathbf{w}$ assigned $-\infty$, so their contribution to $\mathbf{\bar{h}}$ is nullified after softmax normalization (Eq. \ref{eq_att}).
The actor and predictor networks are three-layer MLPs. The classifier network is a 3D CNN with an input of (30, 128, 128) and the same architecture as in \cite{ehj}.
An embedding $\mathbf{h}$ is computed by averaging the CNN encoder outputs over 30-frame arrays extracted from a clip with a sliding window of stride 30.
\section{Experiments} \label{sec:experiments}
\textbf{Data:}
We used a compilation of private echo datasets, divided into training and testing cohorts by ensuring a complete separation in acquisition site between the cohorts. Training data consisted of echo studies from 8 countries across 9 sites, comprising a total of 1720 studies (1382 controls, 338 amyloidosis) and 28,100 clips. The dataset included control cases from patients with aortic stenosis, heart failure (HF), hypertension, hypertrophic obstructive cardiomyopathy, controls confirmed by scintigraphy scans and healthy individuals. The test dataset was collected from 6 countries across 9 sites and contained 1640 studies (1422 controls and 218 amyloidosis) with a total of 33,278 clips. Control cases included HF patients, controls confirmed by scintigraphy or MRI and healthy subjects from the WASE Normals, a large multi-site dataset acquired in 15 countries~\cite{miyoshi2020left}. All data were filtered to retain clips from PLAX, PSAX and A4C views. Clips were preprocessed to remove information outside the ultrasound sector, temporally resampled to 30 frames per second and resized to $128\times128$.
\\
\textbf{Implementation details:}
The RL framework was implemented in PyTorch, trained on 8 parallel environments over 500,000 timesteps on a Nvidia GeForce RTX 3090 Ti. The policy and value networks were optimized using Adam with learning rate annealing. $L2$ normalization was applied to the critic’s optimizer to prevent feature dominance from learned weights $\mathbf{w}_t$. The loss function hyperparameters follow the PPO coefficients $\epsilon=0.2$, with $\lambda_{V}=0.5$, $\lambda_{entropy}=0.01$, $\gamma=0.99$ and $\lambda_{advantage}=0.95$~\cite{schulman2017proximal}. The processing cost was set to $c_t=1$ and the weighting between performance and cost $\lambda_{cost} = 0.05$. $N_{max}$ was set to 200.

We alternate between sampling data from the policy and 4 epochs of optimization per iteration with 4 mini-batches of size 128, updating the policy and value networks. 
Study and clip-level shuffling was applied at each training epoch, to prevent the agent from overfitting to the ordering of clips of each study. For each episode, we initialize $s_0$ by sampling from a Gaussian distribution (mean and variance estimated from training data). Finally, we apply action masking, when appropriate, to avoid 
early convergence to a suboptimal policy. We specifically enforce processing of at least two clips, if available, and restrict view selection if the view has no clips.
\\
\textbf{Benchmarks and metrics:}
We compute sensitivity, specificity, and AUC at a study level, as well as a cost equal to the relative number of clips used compared to the total clips. We compare with the following benchmarks and ablations:
\begin{enumerate}
    \item \textit{All clips}: average of amyloidosis CNN predictions from all clips per study~\cite{holste2022self}.
    \item \textit{Weighted clips:} weighted average of amyloidosis CNN predictions from all clips per study~\cite{wessler2023automated}. The weights are equal to the probabilities that clips are A4C, PLAX, or PSAX view and are obtained from a view classifier\footnote{This was approach was effective in a multi-view aortic stenosis classification~\cite{wessler2023automated}. We trained a view classifier with the same architecture of the amyloidosis CNN.}.
    \item \textit{A4C clips}: average of amyloidosis CNN predictions from A4C clips (most common view). 
    \item \textit{Random sample:} average of amyloidosis CNN predictions from random samples of the three views across all studies. The total samples per view was equal to the number of clips selected by our method. This benchmark evaluates how our method adapts clip selection to each individual study.
    \item \textit{Single clip}: CNN prediction of a randomly selected clip of an arbitrary view.
    \item \textit{STD:} heuristic ``agent'' which keeps processing randomly selected clips if the standard deviation of their amyloidosis predictions is greater than 0.241 (determined to maximize test AUC). The per-study prediction is the mean predictions of selected clips.
    \item \textit{AB1}: ablation of our method where $s_t$  corresponds to CNN predictions and view label (instead of embeddings), and $y_{pred}$  is the average of predictions.
    \item \textit{AB2}: ablation of the proposed method, which does not use attention and the unified embedding $\mathbf{\bar{h}}$ is the average of embeddings $\mathbf{h}_t$.
\end{enumerate}
\section{Results}
\begin{figure}[htb!]
\centering
\includegraphics[width=0.7\textwidth]{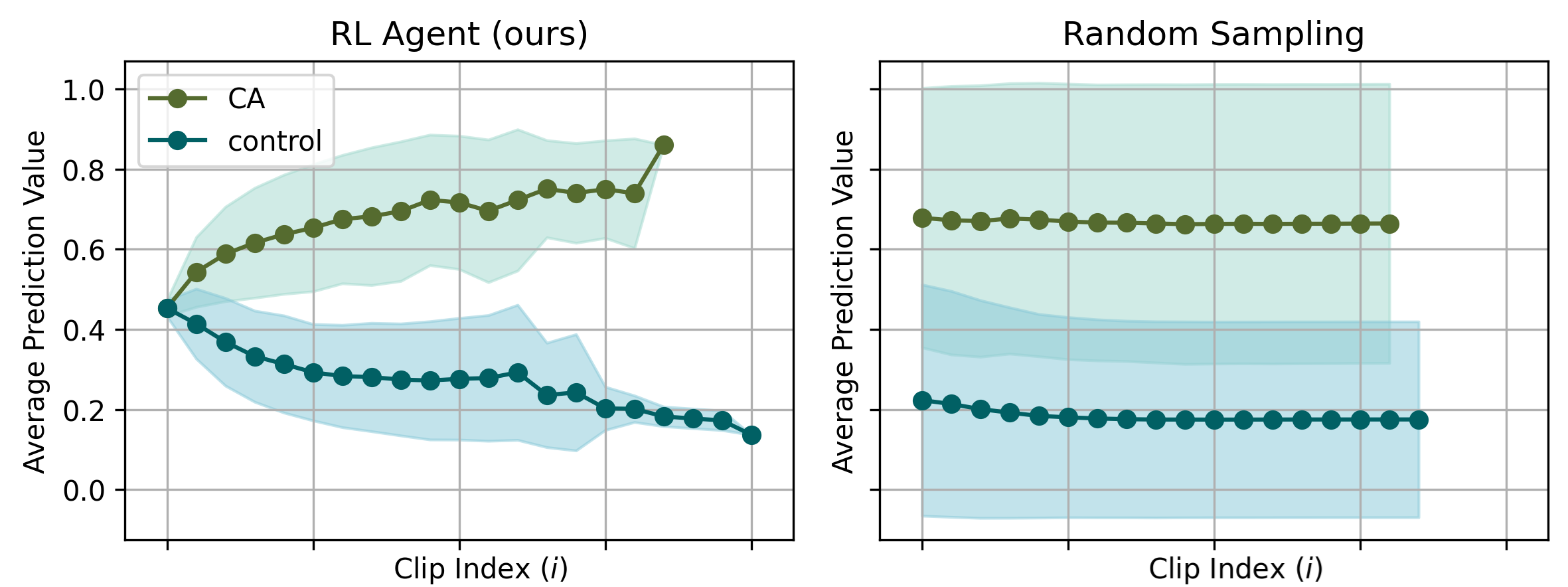}
\caption{Mean prediction for cardiac amyloidosis (green) and control (blue), as clips are sequentially processed by the RL agent (left) and random sampling (right). The shaded area extends one standard deviation, indicating confidence on the mean prediction.} \label{conf_curve}
\end{figure}

\begin{figure}[hbt!]
\centering
\includegraphics[width=0.75\textwidth]{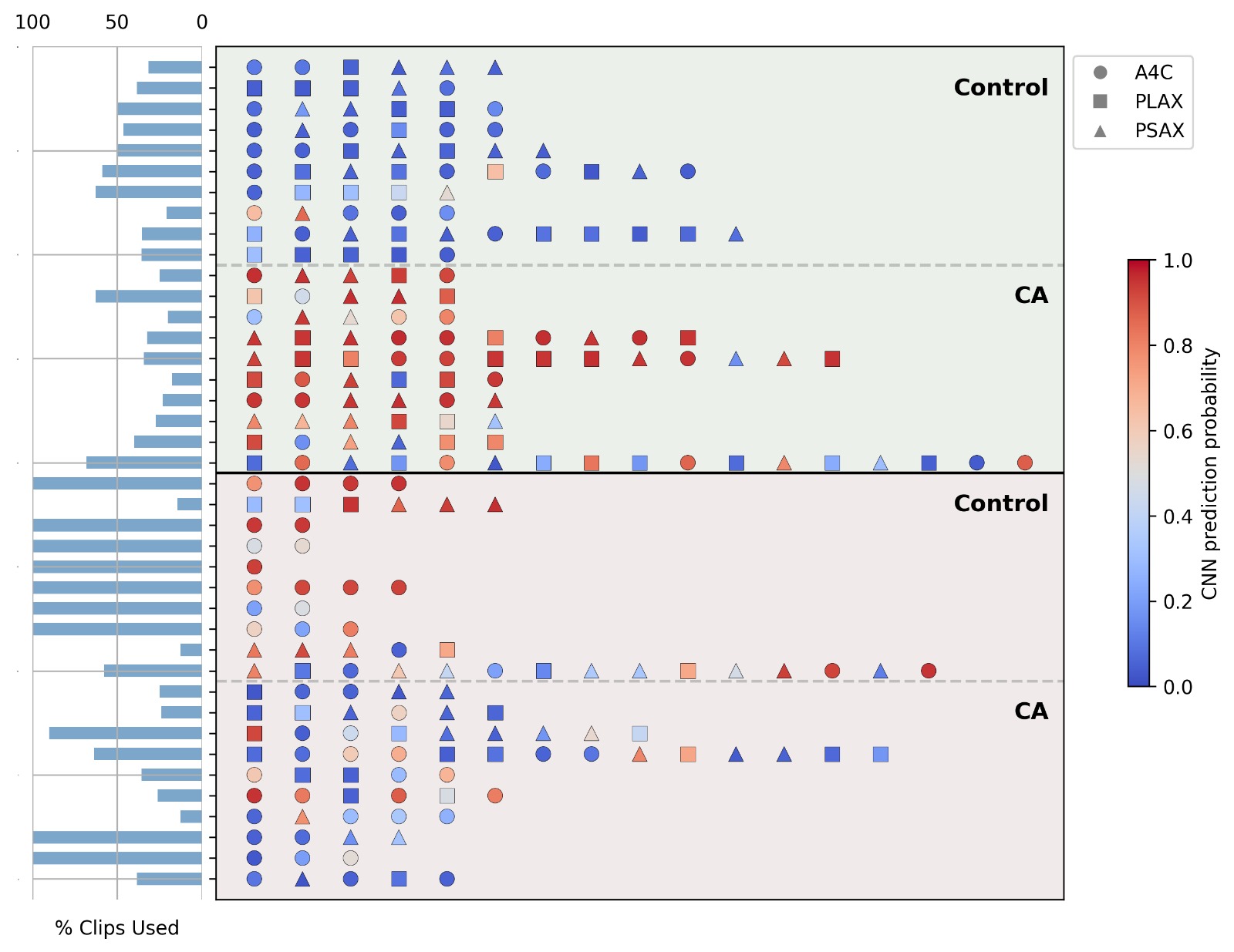}
\caption{RL-based adaptive clip selection. Each row represents a study, with markers indicating the view of each processed clip (\protect\markerone =A4C,
\protect\markertwo=PLAX, \protect\markerthree=PSAX). Color represents the CNN prediction probability (\textcolor{red}{CA}, \textcolor{blue}{control}). Horizontal bars on the left indicate the fraction of clips used per study. Top half (shaded in green) shows correctly classified studies and bottom half (shaded in red) shows misclassified studies.
}
\label{fig:scatter}
\end{figure}

Quantitative results are presented in Table~\ref{tab:performance_comparison}. \textit{Single clip} establishes a lower-bound performance and cost, used to show the benefit of using multiple clips. Averaging the disease predictions from \textit{all clips}, especially after weighting, improves AUC, albeit at the cost of processing all clips.
%
%
%
The cost can be reduced simply by considering all \textit{A4C clips},
or by the heuristic method using the standard deviation to determine how many clips to process. Both have a lower AUC compared to \textit{weighted clips}.
\begin{table}[tb!]
\scriptsize
\centering
\caption{Performance on all, and uncertain studies (subset with highest prediction variability across clips) measured by sensitivity, specificity, $AUC$, and $cost$\% (number of clips used relative to the total). Best results in bold. Values in subscript refer to the standard deviation across 10 runs with different seeds.}
\begin{tabular}{
    l
    @{\hskip 1pt}c@{\hskip -5pt}c@{\hskip -5pt}c@{\hskip -5pt}c
    @{\hskip -5pt}c@{\hskip -5pt}c@{\hskip -5pt}c@{\hskip -5pt}c
}
\toprule
 & \multicolumn{4}{c@{\hskip 4pt}}{All Studies} & \multicolumn{4}{c}{Uncertain Studies} \\
 & $Sens$ & $Spec$ & $AUC$ & $Cost \%$ & $Sens$ & $Spec$ & $AUC$ & $Cost \%$ \\
  \cmidrule(lr){0-0}
 \cmidrule(lr){2-5}
 \cmidrule(lr){6-9}
All clips         & 0.73 & 0.86 & 0.88 & 100.00 & 0.64 & 0.72 & 0.75 & 28.82 \\
Weighted clips    & 0.70 & 0.94 & 0.90 & 100.00 & 0.64 & 0.73 & 0.75 & 28.82 \\
A4C clips         & 0.67 & 0.87 & 0.89 & 26.09  & 0.57 & 0.70 & 0.70 & 7.10  \\
Single clip       & \hspace{0.9em} $0.69_{.03}$ & \hspace{0.9em} $0.82_{.01}$ & \hspace{1.3em} $0.81_{.02}$ & \textbf{4.93} & \hspace{0.9em} $0.56_{.03}$ & \hspace{0.9em} $0.64_{.03}$ & \hspace{1.3em} $0.63_{.03}$ & \textbf{6.14} \\
Random sample     & \hspace{0.9em} $0.73_{.01}$ & \hspace{0.9em} $0.86_{.00}$ & \hspace{1.3em} $0.88_{.01}$ & \hspace{0.9em} $30.02_{1.58}$ & \hspace{0.9em} $0.61_{.02}$ & \hspace{0.9em} $0.71_{.02}$ & \hspace{1.3em} $0.70_{.01}$ & \hspace{1.3em} $8.22_{1.55}$ \\
STD               & \hspace{0.9em} $0.72_{.01}$ & \hspace{0.9em} $0.86_{.00}$ & \hspace{1.3em} $0.87_{.00}$ & \hspace{0.9em} $26.94_{.04}$ & \hspace{0.9em} $0.65_{.01}$ & \hspace{0.9em} $0.72_{.01}$ & \hspace{1.3em} $0.73_{.01}$ & \hspace{1.3em} $12.16_{.53}$ \\
  \cmidrule(lr){0-0}
\cmidrule(lr){2-5}
\cmidrule(lr){6-9}
Ours (AB1)        & \hspace{0.9em} $0.75_{.01}$ & \hspace{0.9em} $0.83_{.01}$ & \hspace{1.3em} $0.86_{.01}$ & \hspace{1.3em} $25.97_{1.58}$ & \hspace{1.3em} $0.66_{.02}$ & \hspace{1.3em} $0.74_{.01}$ & \hspace{1.3em} $0.74_{.01}$ & \hspace{1.3em} $6.88_{1.39}$ \\ 
Ours (AB2)        & \hspace{0.9em} $0.77_{.00}$ & \hspace{0.9em} $0.84_{.01}$ & \hspace{1.3em} $0.89_{.01}$ & \hspace{1.3em} $30.71_{1.72}$ & \hspace{1.3em} $0.69_{.01}$ & \hspace{1.3em} $0.73_{.02}$ & \hspace{1.3em} $0.75_{.01}$ & \hspace{1.3em} $8.37_{1.71}$ \\ 
Ours              & \hspace{1em} $\mathbf{0.78_{.01}}$ & \hspace{1em} $\mathbf{0.87_{.01}}$ & \hspace{1.6em} $\mathbf{0.91_{.01}}$ & \hspace{1.3em} $30.02_{1.58}$ & \hspace{1.3em} $\mathbf{0.73_{.02}}$ & \hspace{1.3em} $\mathbf{0.77_{.01}}$ & \hspace{1.3em} $\mathbf{0.81_{.01}}$ & \hspace{1.3em} $8.22_{1.55}$ \\
\bottomrule
\end{tabular}
\label{tab:performance_comparison}
\end{table}
Random sampling 30\% of the clips can achieve comparable performance to averaging predictions from all clips, which is likely due to high redundancy among clips. The highest performance in sensitivity, specificity, and AUC is achieved by our method, while maintaining a low cost. Our method also outperforms the ablation experiment AB2, demonstrating the importance of using the attention-based embedding pooling instead of the average embedding, and also outperforms AB1 suggesting that embeddings provide a richer source of information for the agent's predictions than using predictions and view labels for the state. When considering only uncertain studies (those with high prediction variability from the disease classifier), the benefit of our method is even larger. 
The performance of our method is attributed to the agent's ability to dynamically decide, for each study, when to stop or when to process more clips from a particular view, unlike the approach of randomly processing clips until reaching a cost level. In cases with high prediction variability, the agent typically selects more clips. As seen in Fig.~\ref{conf_curve}, prediction variability decreases as more clips are selected by the agent, whereas randomly selecting clips maintains a larger prediction spread. When the predictions from the selected clips are consistent, the agent can decide to stop and avoid the cost of additional processing (Fig.~\ref{fig:scatter}). Otherwise, the agent can pick clips from the view that is more likely to complement the information of previously selected clips to further reduce uncertainty. Misclassifications occur in two scenarios: studies with few clips due to lack of information, and studies where discriminative features may not be consistently expressed across all views (Fig.~\ref{fig:scatter}).

\section{Conclusion}
We have proposed a method that selects and leverages information from multiple echo clips in a study, for optimizing disease classification. A sequential decision model optimized through RL learns to either stop when the clips selected are enough to make an accurate prediction, or to process more clips of a particular view to reduce prediction uncertainty. Our method was evaluated on CA diagnosis where it outperformed standard aggregation methods (e.g. all-clip averaging) and a random sampling benchmark. The ability to improve performance by using multiple, but not all clips in a study may facilitate deployment when dealing with large numbers of studies or hardware constraints (e.g. echo devices). 
\newpage
\bibliographystyle{splncs04}
\bibliography{mybibliography.bib}
\end{document}